\documentclass[10pt,conference]{IEEEtran}
\usepackage{comment}
\usepackage{graphicx}
\usepackage{float}
\usepackage{subcaption}
\usepackage{amsmath, amssymb}
\usepackage{algorithm}
\usepackage{algorithmic}
\usepackage{nicematrix, booktabs}
\usepackage{pifont}
\usepackage{multirow}
\usepackage{threeparttable}
\usepackage{xcolor}
\usepackage{makecell}
\usepackage{etoolbox}
\usepackage[backend=biber, style=ieee]{biblatex}
\usepackage[colorlinks=true, allcolors=blue]{hyperref}

\begin{document}

\title{\textsc{Blossom}: Block-wise Federated Learning Over Shared and Sparse Observed Modalities}
\author{
    \IEEEauthorblockN{
        Pranav M R\IEEEauthorrefmark{1}\IEEEauthorrefmark{2},
        Jayant Chandwani\IEEEauthorrefmark{1}\IEEEauthorrefmark{2},
        Ahmed M. Abdelmoniem\IEEEauthorrefmark{3},
        Arnab K. Paul\IEEEauthorrefmark{2}
    }
    \IEEEauthorblockA{\IEEEauthorrefmark{2}\textit{DaSH Lab, BITS Pilani, KK Birla Goa Campus, India} \IEEEauthorrefmark{3}\textit{Queen Mary University of London, United Kingdom}}
    \IEEEauthorblockA{\{f20230340, f20230356, arnabp\}@goa.bits-pilani.ac.in, ahmed.sayed@qmul.ac.uk}
}
\maketitle

\makeatletter
\long\def\@makefntext#1{\noindent #1}
\makeatother

\begingroup
\renewcommand\thefootnote{}
\footnotetext{
    \IEEEauthorrefmark{1} Pranav M R and Jayant Chandwani are the lead authors and contributed equally to this work.
    \vspace{2pt}
    \newline
    This work was supported in part by BITS Pilani CDRF grant C1/23/173, ANRF/SERB SRG grant SRG/2023/002445, and e6Data Inc.
}
\endgroup

\begin{abstract}

Multimodal federated learning (FL) is essential for real-world applications such as autonomous systems and healthcare, where data is distributed across heterogeneous clients with varying and often missing modalities. However, most existing FL approaches assume uniform modality availability, limiting their applicability in practice. We introduce \textsc{Blossom}, a task-agnostic framework for multimodal FL designed to operate under shared and sparsely observed modality conditions. \textsc{Blossom} supports clients with arbitrary modality subsets and enables flexible sharing of model components. To address client and task heterogeneity, we propose a block-wise aggregation strategy that selectively aggregates shared components while keeping task-specific blocks private, enabling partial personalization. We evaluate \textsc{Blossom} on multiple diverse multimodal datasets and analyse the effects of missing modalities and personalization. Our results show that block-wise personalization significantly improves performance, particularly in settings with severe modality sparsity. In modality-incomplete scenarios, \textsc{Blossom} achieves an average performance gain of 18.7\% over full-model aggregation, while in modality-exclusive settings the gain increases to 37.7\%, highlighting the importance of block-wise learning for practical multimodal FL systems.

\end{abstract}

\section{Introduction}

\noindent Federated Learning (FL)~\cite{fedavg} enables collaborative model training across decentralized and privacy-sensitive data sources without sharing raw data. By keeping data local and exchanging only model updates, FL supports learning under strict privacy, legal, and operational constraints, making it well suited for domains such as healthcare~\cite{fl_for_healthcare_survey}, autonomous systems~\cite{fl_for_autonomous_systems}, and mobile sensing~\cite{fl_for_mobile_sensing_survey}.

Many real-world applications are inherently multimodal especially in FL~\cite{multimodal_ml_survey}, relying on heterogeneous information sources such as vision, audio, text, and sensor signals. Leveraging multiple modalities typically yields stronger representations by capturing complementary aspects of the underlying phenomena. In federated settings, however, multimodality is further complicated by client heterogeneity: clients often collect different subsets of modalities due to hardware limitations, cost constraints, or deployment environments~\cite{multimodal_fl_survey_2, harmony}.

Despite this, most FL research primarily addresses statistical heterogeneity (Non-IID data distribution)~\cite{niid_fl_survey} while implicitly assuming uniform modality availability. Even recent multimodal FL approaches frequently rely on full modality overlap or consider only limited forms of missing modalities; such as sample-level corruption or partial observations within a fixed modality set~\cite{multimodal_fl_survey_2}, which does not reflect real-world conditions. In practice, missing modalities arise in several forms: \emph{(i) modality-complete}, where all clients observe all modalities; \emph{(ii) modality-incomplete}, where some modalities are missing for subsets of clients; and \emph{(iii) modality-exclusive}, where clients possess entirely disjoint modality sets. Existing methods typically handle only one of these scenarios, limiting their general applicability.

To address this gap, we propose \textsc{Blossom}, a unified framework for robust multimodal FL under arbitrary modality availability. \textsc{Blossom} decomposes a multimodal model into modality-specific encoders, a fusion module, and task-specific prediction heads. We adopt a late-fusion design and introduce a block-wise aggregation strategy in which only modality encoders shared across clients are aggregated, allowing clients with disjoint modality sets to participate.

To mitigate performance degradation of clients with sparse or exclusive modalities, \textsc{Blossom} incorporates partial personalization~\cite{partial_personalization} by maintaining private task-specific heads per client, enabling shared representations to adapt to local data distributions. We further study variants that also privatize the fusion module, analysing the trade-offs between generalization and personalization under varying modality sparsity.

We evaluate \textsc{Blossom} across diverse multimodal benchmarks spanning emotion recognition, human activity recognition, healthcare, and multimedia tasks. Our results demonstrate consistent improvements over standard FL baselines across modality-complete, incomplete, and exclusive settings, with an average gain of $19.8\%$ across all our experiments. To the best of our knowledge, this work constitutes the first systematic benchmarking study of partial personalization in multimodal FL under missing-modality settings.

To support reproducibility, we release \textsc{Blossom} as an open-source framework with publicly hosted experimental splits (\url{https://github.com/DaSH-Lab-CSIS/blossom}).

\section{Related Work}
\label{sec:related_works}

\begin{figure*}[t]
    \centering
    \includegraphics[width=\textwidth]{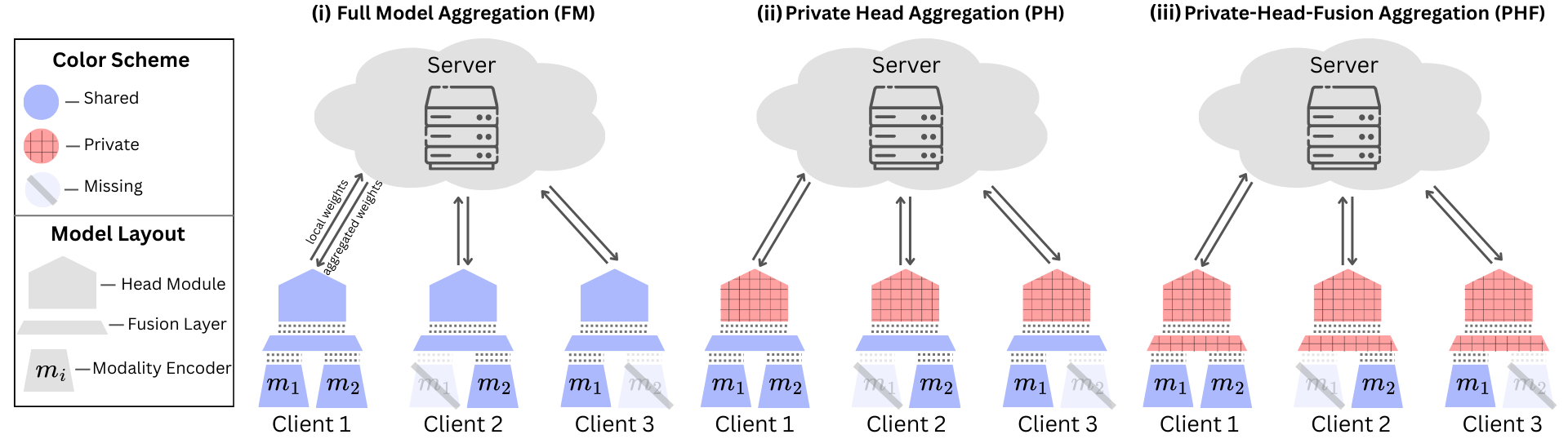}
    \caption{Illustration of the \textsc{Blossom} framework under the three block-wise aggregation modes.}
    \label{fig:architecture}
\end{figure*}

\subsection{Multimodal Learning}
\label{subsec:multimodal_learning}

\noindent Multimodal learning methods are commonly categorized into early and late fusion strategies~\cite{multimodal_ml_survey}. Early fusion combines low-level features from all modalities into a joint representation and requires all modalities to be present during training and inference. Late fusion processes each modality independently and integrates high-level representations at a later stage, making it more robust to missing-modality scenarios~\cite{multimodal_fl_survey_1}.

These distinctions are critical in federated settings with heterogeneous modality availability across clients. Standard Federated Averaging (FedAvg)~\cite{fedavg} assumes homogeneous model architectures and synchronized parameter updates, and therefore performs poorly under missing or asymmetric modality distributions. Multimodal FL addresses this by extending FL to clients observing different modality subsets, but introduces challenges such as incompatible parameter updates, restricted aggregation, and performance degradation under sparse or disjoint modality observations.

\subsection{Multimodal FL with Sparse Modalities}
\label{subsec:multimodal_fl_with_sparse_modalities}

\noindent Early multimodal FL approaches largely avoid explicit multimodal modeling under sparsity by decoupling modalities. Ensemble-based methods, such as FedMFS~\cite{fedmfs_ensemble}, train independent unimodal models and combine predictions at inference time. While robust to missing modalities, they cannot capture cross-modal interactions and incur high communication overhead. KD-based approaches, exemplified by CreamFL~\cite{creamfl}, enable cross-modal transfer via distillation but typically require a public dataset at the server, weakening federated privacy guarantees. Both classes implicitly assume \emph{modality sufficiency}, that each modality alone contains enough information to solve the task, an assumption inspired by classical multi-view learning~\cite{co_training_modality_sufficient} that often fails under sparse or heterogeneous modality availability.

Representation learning approaches move beyond unimodal decoupling by learning modality-specific latent embeddings, commonly via autoencoders~\cite{iot}, which are later fused for prediction. While this enables limited cross-modal interaction, recent studies show that autoencoder-based representations remain vulnerable to gradient-based reconstruction attacks~\cite{autoencoder_vulnerability}, raising privacy concerns in federated settings.

Fusion-based methods explicitly separate modality encoders from task-level components and combine them via late fusion, making them more suitable for heterogeneous federated environments. Hybrid variants, such as Harmony~\cite{harmony}, integrate ensemble-style training with fusion-based aggregation but rely on sufficient cross-client modality overlap and degrade under highly sparse regimes. FedMSplit~\cite{fedmsplit} more directly targets missing-modality settings using graph-based aggregation over overlapping modality subsets; however, its reliance on overlap limits applicability in modality-exclusive scenarios.

Unlike prior work such as FedMultimodal~\cite{fedmultimodal} and FedMSplit, we focus on \emph{structural modality missingness}, where clients lack specific modalities entirely rather than individual samples being partially observed. This client-level modality heterogeneity is formalized in Section~\ref{sec:evaluation}. Very recent work also considers fully missing client modalities~\cite{fl_with_fully_missing}, but most such methods rely on reconstructing absent modalities, which conflicts with federated privacy constraints.

\subsection{Partial Personalization}

\noindent Partial personalization in FL refers to the joint training of shared global parameters and client-specific local parameters, where only the shared parameters are aggregated across clients. In multimodal settings, personalization can be applied at either the encoder or task-head level. While encoder-level personalization improves local adaptation, it restricts cross-client representation sharing, which is undesirable when clients observe sparse or disjoint modality subsets. Task-head-level personalization instead enables clients to adapt shared multimodal representations to local label distributions while preserving collaborative representation learning.

Prior work on partial personalization analyses how separating shared and private model components improves optimization stability, convergence, and generalization under heterogeneous client data distributions~\cite{partial_personalization}. However, this line of work does not consider heterogeneity arising from missing or disjoint modalities across clients.

\section{Methodology}

\subsection{Problem Formulation}
\noindent We consider a multimodal FL setting with $\mathcal{C}$ clients and $\mathcal{M}$ distinct modalities. Each client $k$ observes data from a subset of modalities $M_k \subseteq \mathcal{M}$, and these subsets may vary arbitrarily across clients. We focus on extremely sparse modality settings, where client modality subsets have little or no overlap. We additionally consider modality-insufficient scenarios, in which no single modality contains sufficient information to solve the task in isolation.

\subsection{Architecture of \textsc{Blossom}}
\noindent We use a late-fusion multimodal architecture as discussed in Section~\ref{subsec:multimodal_learning} where each modality is encoded separately and combined only at the fusion stage.

As shown in Figure~\ref{fig:architecture}, \textsc{Blossom} is organized into three blocks: modality-specific encoders, a fusion module, and a task prediction head. Encoders produce latent modality representations, the fusion block integrates the representations from the modalities available at a client, and the head generates the final output. Missing modalities are handled by zeroing out the corresponding encoder outputs before fusion, as proposed by \textcite{partial_personalization}, preserving a fixed fusion architecture and input dimensionality. This block structure enables block-wise aggregation and partial personalization.

Within this setup, we study two fusion methods: concatenation-based fusion (ConcatFusion), which merges modality embeddings by concatenation followed by a projection layer, and attention-based fusion (AttentionFusion), which learns modality-dependent weights to combine embeddings adaptively. Each fusion method is evaluated under two partial-personalization configurations to compare robustness under sparse and heterogeneous modality availability.

\subsection{Block-wise Federated Aggregation}

\noindent \textbf{Aggregation Backbone:} We adopt FedAvg as the underlying aggregation primitive. It remains the most widely used and well-understood aggregation strategy in FL and serves as a common baseline in multimodal FL studies. Furthermore, empirical findings reported in FedMultimodal~\cite{fedmultimodal} indicate that alternative federated optimizers (except FedOpt) yield only marginal performance differences under comparable architectures and data partitions.

This choice allows us to isolate and analyse the impact of block-wise aggregation and partial personalization without introducing confounding effects from more complex server-side optimization schemes. Importantly, \textsc{Blossom} is \emph{aggregation-strategy agnostic}: the proposed block-wise decomposition and selective aggregation mechanism can be readily combined with alternative federated optimizers (e.g., FedAdam, FedYogi~\cite{fedadam}) without requiring changes to the model architecture or training protocol.

\noindent \textbf{Block-wise Aggregation:} We implement block-wise federated aggregation through the server-side procedure specified in Algorithm~\ref{alg:aggregation_strategy}. Instead of aggregating model parameters as a single unit, the server aggregates each architectural block separately according to block type and modality availability. In particular, modality encoders are aggregated only from clients that possess the corresponding modality (lines 11–14), while the fusion and head blocks are either aggregated or kept client-private depending on the selected personalization mode (lines 15–25). This enables training with missing modalities and uneven modality overlap across clients.

\begin{algorithm}[t]
\small
\caption{Block-wise Aggregation Strategy.}
\label{alg:aggregation_strategy}
\begin{algorithmic}[1]

\STATE \textbf{Input:} Clients $\mathcal{C}$, modalities $\mathcal{M}$, aggregation method $\mathcal{A}(\cdot)$
\STATE \textbf{Hyperparameters:} $T$, $E$, aggregation mode $\mathcal{P} \in \{\texttt{FM}, \texttt{PH}, \texttt{PHF}\}$

\STATE \textbf{Initialize:} Global modality encoders $\{\theta_m\}_{m \in \mathcal{M}}$, fusion block $\theta_f$, head block $\theta_h$

\FOR{each communication round $t = 1$ to $T$}
    \STATE Server sends relevant global blocks to participating clients
    \FOR{each client $c \in \mathcal{C}$ \textbf{in parallel}}
        \STATE Initialize local model using received blocks
        \STATE Train local model on client data for $E$ epochs
        \STATE Send updated blocks to server
    \ENDFOR

    \FOR{each modality $m \in \mathcal{M}$}
        \STATE Get encoder updates $\{\theta_m^c\}$ from clients with modality $m$
        \STATE $\theta_m \leftarrow \mathcal{A}(\{\theta_m^c\})$
    \ENDFOR

    \IF{$\mathcal{P} == \texttt{FM}$}
        \STATE Collect fusion updates $\{\theta_f^c\}$ and head updates $\{\theta_h^c\}$
        \STATE $\theta_f \leftarrow \mathcal{A}(\{\theta_f^c\})$
        \STATE $\theta_h \leftarrow \mathcal{A}(\{\theta_h^c\})$
        
    \ELSIF{$\mathcal{P} == \texttt{PH}$}
        \STATE Collect fusion updates $\{\theta_f^c\}$
        \STATE $\theta_f \leftarrow \mathcal{A}(\{\theta_f^c\})$
        \STATE \textbf{Keep} $\theta_h$ private to each client
        
    \ELSIF{$\mathcal{P} == \texttt{PHF}$}
        \STATE \textbf{Keep} $\theta_f$ and $\theta_h$ private to each client
    \ENDIF
\ENDFOR

\RETURN updated global modality encoders $\{\theta_m\}_{m \in \mathcal{M}}$
\end{algorithmic}
\end{algorithm}

\noindent \textbf{Personalization Modes:} The personalization behavior is controlled by the aggregation mode parameter $\mathcal{P}$ in Algorithm~\ref{alg:aggregation_strategy}. We evaluate three configurations:

\begin{enumerate}
    \item \emph{Full model aggregation (FM):} all blocks are aggregated (lines 15–18); this matches the standard aggregation setting used in most prior multimodal FL works and serves as our primary reference configuration.
    \item \emph{Private head (PH):} modality encoders and the fusion block are aggregated, while the prediction head remains client-specific (lines 19–21); we denote this head-personalized configuration as PH for clarity. Prior work has explored comparable personalization patterns, though not under consistent naming.
    \item \emph{Private head-fusion (PHF):} only modality encoders are aggregated, while clients keep both the fusion and head blocks private (lines 22–24); we refer to this encoder-only aggregation regime as PHF, again as a naming convenience rather than a new personalization concept.
\end{enumerate}

\begin{table*}[t]
\footnotesize
\centering
\caption{Overview of multimodal datasets used to evaluate \textsc{Blossom}.}
\label{tab:datasets}
\resizebox{\textwidth}{!}{
\begin{NiceTabular}{llcccc}
\toprule
\textbf{Task} 
& \textbf{Datasets}
& \textbf{Modalities}
& \textbf{Features}
& \textbf{Models}
& \textbf{Metric} \\
\midrule
HAR
& KU-HAR~\cite{kuhar}, UCI-HAR~\cite{ucihar}
& Acc, Gyro
& Raw, Raw
& Conv1D+GRU, Conv1D+GRU
& F1
\\
Healthcare
& PTB-XL~\cite{ptbxl}
& I-AVF, V1-V6
& Raw, Raw
& Conv1D+GRU, Conv1D+GRU
& F1
\\
Multimedia
& AV-MNIST~\cite{avmnist}
& Image, Audio
& Raw, MelSpec
& Conv2D, Conv2D
& Accuracy
\\
Emotion Recognition
& MELD~\cite{meld}, IEMOCAP~\cite{iemocap}
& Audio, Text
& MelSpec, MobileBERT
& Conv1D+GRU, GRU
& Accuracy \\
\bottomrule
\end{NiceTabular}
}
\end{table*}

\begin{table*}[t]
\scriptsize
\centering
\caption{Performance comparison of different aggregation modes.}
\label{tab:results}

\resizebox{\textwidth}{!}{
\begin{NiceTabular}{| l | c | ccc ccc | ccc ccc |}
\toprule
\multirow{3}{*}{\textbf{Dataset}}
& \multirow{3}{*}{\makecell{\textbf{Modality}\\\textbf{Config}}}
& \multicolumn{6}{c}{\textbf{ConcatFusion}}
& \multicolumn{6}{c}{\textbf{AttentionFusion}} \\

\cmidrule(lr){3-8} \cmidrule(lr){9-14}

& 
& \multicolumn{3}{c}{\textbf{IID}}
& \multicolumn{3}{c}{\textbf{NIID}}
& \multicolumn{3}{c}{\textbf{IID}}
& \multicolumn{3}{c}{\textbf{NIID}} \\

\cmidrule(lr){3-5} \cmidrule(lr){6-8}
\cmidrule(lr){9-11} \cmidrule(lr){12-14}

& 
& \textbf{FM}
& \textbf{PH}
& \textbf{PHF}
& \textbf{FM}
& \textbf{PH}
& \textbf{PHF}
& \textbf{FM}
& \textbf{PH}
& \textbf{PHF}
& \textbf{FM}
& \textbf{PH}
& \textbf{PHF} \\
\midrule

\multirow{3}{*}{KU-HAR}
& 0--0--10
& \textbf{90.14} & 83.39 & 84.86 & \textbf{66.91} & 64.96 & 64.58
& \textbf{86.61} & 79.98 & 73.95 & \textbf{63.48} & 62.22 & 61.41 \\
& 3--3--4
& 71.11 & \textbf{74.28} & 73.30 & 45.39 & 57.15 & \textbf{58.77}
& 63.06 & 62.36 & \textbf{65.30} & 46.16 & 49.31 & \textbf{53.46} \\
& 5--5--0
& 59.29 & \textbf{73.52} & 71.49 & 35.28 & \textbf{55.96} & 54.10
& 31.03 & 51.77 & \textbf{64.64} & 21.91 & 33.06 & \textbf{50.29} \\

\midrule
\multirow{3}{*}{UCI-HAR}
& 0--0--10
& \textbf{91.10} & 87.45 & 88.60 & \textbf{83.95} & 80.79 & 82.54
& \textbf{87.65} & 85.20 & 84.72 & \textbf{80.78} & 80.16 & 76.50 \\
& 3--3--4
& 71.11 & \textbf{74.28} & 73.30 & 45.39 & 57.15 & \textbf{58.77}
& \textbf{80.38} & 75.02 & 78.15 & \textbf{71.69} & 67.68 & 71.10 \\
& 5--5--0
& 59.29 & \textbf{73.52} & 71.49 & 35.28 & \textbf{55.96} & 54.10
& 79.79 & 70.95 & \textbf{80.17} & 69.41 & \textbf{70.80} & 66.78 \\

\midrule
\multirow{3}{*}{PTB-XL}
& 0--0--10
& \textbf{66.13} & 63.59 & 63.78 & \textbf{55.91} & 55.31 & 55.76
& \textbf{64.32} & 63.41 & 60.77 & 54.25 & \textbf{55.11} & 53.83 \\
& 3--3--4
& 41.24 & 59.90 & \textbf{60.20} & 38.97 & 51.60 & \textbf{52.10}
& 37.10 & 42.18 & \textbf{56.66} & 35.16 & 41.71 & \textbf{50.84} \\
& 5--5--0
& 24.49 & 58.17 & \textbf{58.19} & 28.28 & \textbf{49.95} & 49.51
& 36.80 & 44.14 & \textbf{55.47} & 38.71 & 34.10 & \textbf{48.79} \\

\midrule
\multirow{3}{*}{AV-MNIST}
& 0--0--10
& \textbf{99.83} & 99.78 & 99.68 & \textbf{99.75} & 99.57 & 99.42
& \textbf{99.81} & 99.45 & 98.58 & \textbf{99.58} & 99.08 & 98.58 \\
& 3--3--4
& 82.34 & 96.26 & \textbf{98.77} & 82.92 & 96.48 & \textbf{98.41}
& 84.45 & 96.36 & \textbf{98.41} & 94.31 & \textbf{97.58} & 97.05 \\
& 5--5--0
& 84.94 & 97.18 & \textbf{98.43} & 78.97 & 97.18 & \textbf{97.88}
& 81.08 & 94.14 & \textbf{98.12} & 92.98 & 78.58 & \textbf{96.19} \\

\midrule
\multirow{3}{*}{MELD}
& 0--0--10
& 61.09 & \textbf{61.68} & 61.55 & 62.28 & 81.18 & \textbf{82.21}
& 59.66 & 61.10 & \textbf{61.60} & 59.91 & \textbf{82.07} & 81.47 \\
& 3--3--4
& 55.93 & 57.62 & \textbf{57.75} & 59.97 & 80.38 & \textbf{81.33}
& 57.48 & 57.02 & \textbf{58.09} & 54.95 & 81.29 & \textbf{81.50} \\
& 5--5--0
& \textbf{55.32} & 55.25 & 54.69 & 56.41 & \textbf{81.03} & 79.80
& 53.94 & 54.72 & \textbf{55.23} & 59.69 & 80.83 & \textbf{80.97} \\

\midrule
\multirow{3}{*}{IEMOCAP}
& 0--0--10
& \textbf{67.09} & 64.55 & 63.92 & 64.38 & 75.82 & \textbf{75.93}
& \textbf{67.86} & 64.26 & 60.34 & 61.61 & \textbf{73.79} & 72.77 \\
& 3--3--4
& \textbf{56.72} & 54.68 & 55.65 & 56.18 & \textbf{71.89} & 70.72
& \textbf{56.75} & 54.42 & 52.95 & 59.96 & 69.78 & \textbf{70.86} \\
& 5--5--0
& \textbf{50.68} & 49.69 & 50.63 & 51.41 & 69.90 & \textbf{70.19}
& \textbf{52.56} & 50.18 & 49.68 & 53.17 & 67.92 & \textbf{69.39} \\

\bottomrule
\end{NiceTabular}
}
\end{table*}

\section{Experimental Setup}
\label{sec:evaluation}

\noindent Our framework is built on top of the Flower FL framework~\cite{flower}, with Hydra-based configuration management~\cite{hydra}, enabling flexible specification of datasets, tasks, modality availability patterns, and aggregation strategies. We evaluate \textsc{Blossom} on a diverse suite of real-world multimodal tasks spanning human activity recognition, healthcare, multimedia classification, and emotion recognition, summarized in Table~\ref{tab:datasets}.

We conduct all experiments using 10 clients, 60 communication rounds, and 1 local epoch per round. For all non-IID (NIID) experiments, we use Dirichlet-based label partitioning ($\alpha=0.5$). We adopt a modality configuration notation of the form a–b–c, where a, b, and c denote the number of clients possessing modality~1 only, modality~2 only, and both modalities, respectively. For example, 0–0–10 corresponds to a modality-complete setting where all clients observe both modalities, while configurations such as 3–3–4 and 5–5–0 introduce increasing levels of missing modalities, analogous to missing-modality rates commonly used in prior work.

Since we focus on structural missing modalities, each absent modality at a client is counted as a missing instance. Under this scheme, 0–0–10 corresponds to 0\% missing modality rate, 3–3–4 to 30\%, and 5–5–0 to 50\%. This setting is substantially more challenging and realistic for FL because clients lack entire modalities, rather than the commonly studied sample-level missingness where clients still partially observe all modalities.

\section{Evaluation and Results}

\noindent The evaluation of \textsc{Blossom} addresses the following research questions:
\begin{itemize}
    \item To what extent does partial personalization improve performance under (1) missing-modality conditions and (2) client-side label heterogeneity? (Section~\ref{subsec:results_1})
    \item Between the PH and PHF personalization modes, which configuration yields better performance across varying levels of modality sparsity? (Section~\ref{subsec:results_2})
    \item How does modality insufficiency impact performance across different tasks? (Section~\ref{subsec:results_3})
    \item Do modality-incomplete clients contribute positively to the final global model performance? (Section~\ref{subsec:results_4})
    \item How does \textsc{Blossom} compare to state-of-the-art multimodal FL benchmarks? (Section~\ref{subsec:results_5})
\end{itemize}
\subsection{Impact of Partial Personalization}
\label{subsec:results_1}

\noindent To quantify the impact of personalization, we define:
{\small
    \[
    \text{PH Gain} = \frac{S_{\text{PH}} - S_{\text{FM}}}{S_{\text{FM}}} \times 100\%,
    \]
}
{\small
    \[
    \text{PHF Gain} = \frac{S_{\text{PHF}} - S_{\text{FM}}}{S_{\text{FM}}} \times 100\%,
    \]
}
{\small
    \[
    \text{Personalization Gain (PG)} = \max(\text{PH Gain}, \text{PHF Gain}),
    \]
}
where \(S\) denotes the validation score of interest.

In the modality-complete setting (0--0--10), FM slightly outperforms the personalized variants, with marginally negative gains ($PG > -5\%$). This outcome follows naturally because 0--0--10 reduces to a standard FL setup analogous to FedAvg, where structural heterogeneity is absent and personalization offers limited benefit. Performance in this regime aligns with established benchmarks for KU-HAR, UCI-HAR, PTB-XL, and MELD~\parencite{fedmultimodal}, and for IEMOCAP~\parencite{iemocap_benchmark}, validating our setup.

\subsubsection{Missing Modalities}
Under modality-incomplete (3--3--4) and modality-exclusive (5--5--0) settings, partial personalization yields substantial performance improvements. Both PH and PHF consistently outperform FM, with an average PG of $19.77\%$. The gain increases from $18.65\%$ in the 3--3--4 case to $37.70\%$ in the 5--5--0 setting, indicating that personalization benefits grow with increasing modality sparsity.

\subsubsection{Label Heterogeneity}
Personalization gains further increase under label heterogeneity. In NIID settings, the average PG reaches $25.82\%$, compared to $13.71\%$ in IID. The largest gains occur when both factors occur simultaneously, with PG of $25.32\%$ for NIID 3--3--4 and $43.28\%$ for NIID 5--5--0, highlighting the robustness of partial personalization under combined structural and distributional heterogeneity. 

\noindent \textbf{Emotion Tasks:} For MELD and IEMOCAP, we observe large personalization gains under NIID splits even in modality-complete settings, likely due to strong label imbalance and emotional ambiguity that make personalized heads better suited for sparse local label distributions. Conversely, these are the only datasets where personalization causes performance drops under IID settings. We hypothesize that for such complex tasks, access to the full global data distribution is more important than local adaptation, which becomes sample-inefficient under uniform data. For other tasks, block-wise aggregation mitigates this IID penalty.

\subsection{PH vs. PHF}
\label{subsec:results_2}

\noindent We compare the two personalization modes, PH and PHF. Across all experiments,
we observe an average PH Gain of $14.85\%$, while PHF Gain is higher at $18.82\%$, indicating a consistent advantage of personalizing the fusion module in addition to the task head.

This difference is fusion-dependent. As shown in Table~\ref{tab:results}, under ConcatFusion, PH Gain and PHF Gain are nearly identical, with average gains of $20.82\%$ and $20.87\%$, respectively. In contrast, under AttentionFusion, PHF Gain substantially exceeds PH Gain, achieving an average gain of $16.76\%$ compared to $8.89\%$ for PH.

This behaviour arises from the nature of the fusion operator. ConcatFusion is a fixed operation, so personalizing it provides little improvement in gain. AttentionFusion, however, is a learned module, and keeping it personalized allows clients to adapt the fusion process to their local modality availability and data distribution, leading to higher gains under both missing-modality and label-heterogeneous settings. Consequently, since attention-based fusion is widely used in practice and PHF Gain consistently matches or exceeds PH Gain, PHF represents a robust default choice for multimodal FL.

\subsection{Modality-insufficient tasks}
\label{subsec:results_3}

\noindent Some datasets in our evaluation are inherently modality-insufficient. PTB-XL is a representative case, as it consists of 12 ECG leads split into two sets of six treated as separate modalities. A single set captures only partial cardiac information, leading to severe performance degradation under missing-modality settings. KU-HAR exhibits a similar trend. As reported in FedMultimodal, performance drops by up to $43\%$ under a 50\% missing modality rate, indicating that no single sensor stream is sufficient for reliable prediction.

Accordingly, we observe the largest personalization gains on these datasets, with average PG of $80.12\%$ for KU-HAR and $72.75\%$ for PTB-XL. While PH and PHF perform similarly under ConcatFusion, under AttentionFusion PHF significantly outperforms PH, achieving gains of $41.47\%$ on KU-HAR and $30.78\%$ on PTB-XL in IID settings.

These results indicate that for modality-insufficient tasks, personalizing the fusion module is particularly critical, as it helps mitigate the compounded effects of limited representational capacity and client heterogeneity.

\subsection{Impact of Modality-Incomplete Clients}
\vspace{-7pt}
\label{subsec:results_4}
\begin{figure}[H]
    \includegraphics[width=\columnwidth]{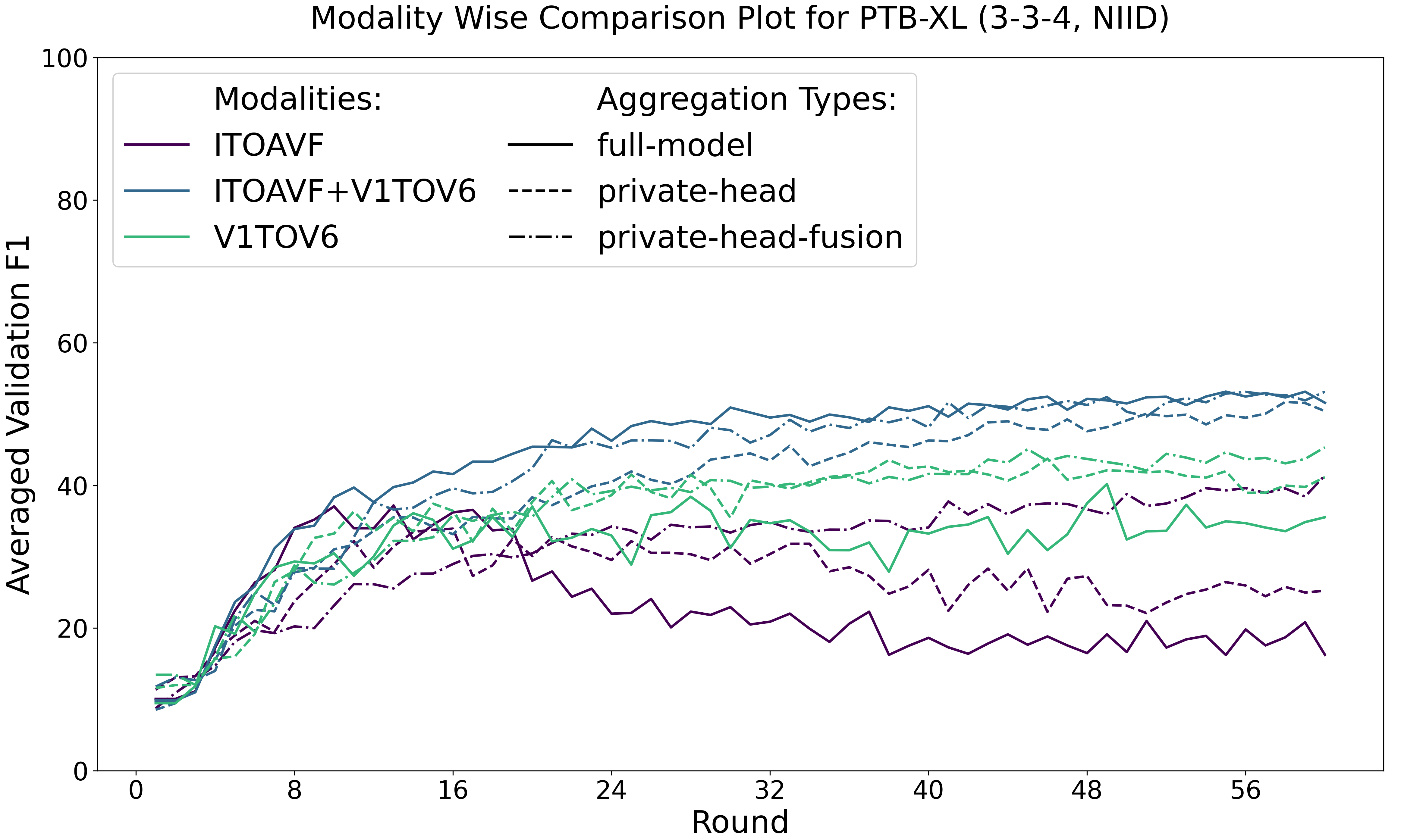}
    \caption{\small Modality-wise validation F1 across training rounds on PTB-XL under the 3–3–4 NIID setting. Colours indicate client modality availability (unimodal vs. multimodal), and line styles denote aggregation modes (FM, PH, PHF).}
    \label{fig:modality_wise_comp}
\end{figure}
\vspace{-7pt}

\noindent An important question in structurally sparse multimodal federated settings is whether modality-incomplete clients meaningfully contribute to global performance. We investigate this using modality-wise validation F1 across training rounds on PTB-XL under the 3–3–4 client configuration, grouped by modality availability and aggregation strategy (Figure~\ref{fig:modality_wise_comp}).

PTB-XL is modality-insufficient by design, so performance depends on effective cross-client modality sharing. We highlight NIID results because the effect is easiest to observe here; however, the same ordering holds across most datasets under both IID and NIID splits, with smaller gaps in the IID case.

Performance differences are most pronounced for unimodal clients, where aggregation choice has the strongest impact. PHF consistently outperforms PH across rounds and often approaches FM scores for multimodal clients, with the largest gains visible for ITOAVF-only clients. In contrast, multimodal clients remain relatively stable during training, with only minor differences between FM, PH, and PHF. Overall, this indicates that personalization substantially improves the usefulness of updates from modality-incomplete clients.

\begin{table}[t]
\footnotesize
\centering
\caption{\small Relative performance degradation under varying missing modality rates (IID, AttentionFusion). Lower is better.}
\label{tab:sota_comparison}

\begin{NiceTabular}{l c cc}
\toprule
\textbf{Dataset} & \makecell{\textbf{Modality Config}\\\textbf{(Missing Rate)}} & \textbf{FedMultimodal} & \textbf{\textsc{Blossom} (PHF)} \\
\midrule
\multirow{2}{*}{KU-HAR}  
& 3--3--4 (30\%) & 13.0\% & \textbf{11.7\%} \\
& 5--5--0 (50\%) & 43.0\% & \textbf{12.6\%} \\

\midrule
\multirow{2}{*}{UCI-HAR} 
& 3--3--4 (30\%) & \textbf{4.6\%} & 7.8\% \\
& 5--5--0 (50\%) & 10.3\% & \textbf{5.4\%} \\

\midrule
\multirow{2}{*}{PTB-XL}  
& 3--3--4 (30\%) & 14.0\% & \textbf{6.8\%} \\
& 5--5--0 (50\%) & 18.0\% & \textbf{8.7\%} \\

\midrule
\multirow{2}{*}{MELD}    
& 3--3--4 (30\%) & \textbf{4.0\%} & 5.7\% \\
& 5--5--0 (50\%) & \textbf{6.7\%} & 10.3\% \\

\bottomrule
\end{NiceTabular}
\end{table}

\subsection{Comparison with FedMultimodal}
\label{subsec:results_5}

\noindent We compare \textsc{Blossom} against the state-of-the-art multimodal FL benchmark FedMultimodal on the four common datasets (KU-HAR, UCI-HAR, PTB-XL, and MELD). AV-MNIST and IEMOCAP are excluded, as they are not part of FedMultimodal, and are used only for internal comparisons.

As noted in Section~\ref{subsec:multimodal_fl_with_sparse_modalities}, most multimodal FL methods, including FedMultimodal, do not explicitly study settings with fully missing client modalities.
In addition, FedMultimodal also uses a different partitioning scheme, which makes direct comparisons of absolute performance unreliable. We therefore report relative percentage performance losses under missing-modality scenarios using corresponding missing rates from FedMultimodal (Section~\ref{sec:evaluation}) in Table~\ref{tab:sota_comparison}.

For consistency with the original FedMultimodal setup, we restrict this comparison to the IID partition and the AttentionFusion configuration, which most closely matches their experimental setup. As shown in Table~\ref{tab:sota_comparison}, despite operating under a more challenging structural-missingness setting, our framework achieves lower relative degradation in most cases compared to FedMultimodal, indicating stronger robustness to modality sparsity.

\section{Conclusion}

\noindent We introduced \textsc{Blossom}, a framework that addresses modality scarcity and exclusivity in FL through a modular late-fusion architecture and block-wise aggregation strategy. Our evaluation confirms that partial personalization crucially mitigates performance degradation in modality-exclusive scenarios while balancing global knowledge sharing.

Despite its effectiveness, our study has limitations that guide future work. Computational constraints restricted experiments to specific modality configurations and client counts, leaving broader scalability and low-participation scenarios unexplored. Additionally, while we evaluated diverse domains, our datasets contain few modalities per client, necessitating validation in richer multimodal settings.

To facilitate further research, we release \textsc{Blossom} as an open-source framework. Future work will leverage the framework's modularity to explore heterogeneous task heads for multi-task learning and extend \textsc{Blossom} to decentralized cross-silo environments.

\printbibliography

\end{document}